\def\year{2022}\relax
\documentclass[letterpaper]{article} 
\usepackage{aaai22}  
\usepackage{times}  
\usepackage{helvet}  
\usepackage{courier}  
\usepackage[hyphens]{url}  
\usepackage{graphicx} 
\usepackage{natbib}  
\usepackage{caption} 
\DeclareCaptionStyle{ruled}{labelfont=normalfont,labelsep=colon,strut=off} 
\usepackage{algorithm}
\usepackage{algorithmic}

\usepackage{newfloat}
\usepackage{listings}
\floatstyle{ruled}
\newfloat{listing}{tb}{lst}{}
\floatname{listing}{Listing}
\title{ChartParser: Automatic Chart Parsing for Print-Impaired}

\author {
    Anukriti Kumar,\textsuperscript{\rm 1}
    Tanuja Ganu, \textsuperscript{\rm 1}
    Saikat Guha \textsuperscript{\rm 1}
}
\affiliations {
    \textsuperscript{\rm 1} Microsoft Research \\

    t-anukumar@microsoft.com, tanuja.ganu@microsoft.com, saikat@microsoft.com
}

\usepackage{bibentry}

\begin{document}

\maketitle

\begin{abstract}
Infographics are often an integral component of scientific documents for reporting
qualitative or quantitative findings as they make it much simpler to comprehend the underlying complex information. However, their interpretation continues to be a challenge for the blind, low-vision, and other print-impaired (BLV) individuals. In this paper, we propose ChartParser, a fully automated pipeline that leverages deep learning, OCR, and image processing techniques to extract all figures from a research paper, classify them into various chart categories (bar chart, line chart, etc.) and obtain relevant information from them, specifically bar charts (including horizontal, vertical, stacked horizontal and stacked vertical charts) which already have several exciting challenges. Finally, we present the retrieved content in a tabular format that is screen-reader friendly and accessible to the BLV users. We present a thorough evaluation of our approach by applying our pipeline to sample real-world annotated bar charts from research papers.

\end{abstract}

\section{Introduction}
Academic research is advancing at an incredible pace, with thousands of scientific documents published monthly \cite{link1}. These documents often use figures/charts as a medium for data representation and interpretation. However, the blind, low-vision and other print-disabled (BLV) individuals are often deprived of insights and understanding offered by these figures. Although these are converted into non-visual, screen-reader friendly representations such as alt-text, data table, etc., there is a lot of reliance on volunteers for this conversion, making it an extremely time-consuming process. In most cases, even the alternate text fails to describe charts properly. Hence, our goal in this paper is to design a fully automated pipeline to extract useful information from charts, specifically bar charts, and convert them into accessible data tables. Potential applications of our system include helping authors provide meaningful captions to their figures in papers, improving search and retrieval of relevant information in the academic domain, generating summaries from charts, building query-answering systems, developing interfaces that can provide simple and convenient access to complex information, making charts accessible for BLV individuals, and helping academic committees and publishers identify plagiarized articles.

Given the remarkable progress in analyzing natural scene images observed in recent years, it is generally assumed that analyzing scientific figures is a trivial task. However, understanding charts/infographics present a plethora of complex challenges. Firstly, a high level of accuracy is expected while parsing the figure plot data, as even a small mistake in analyzing chart data can lead to erroneous conclusions. Also, authors employ different design conventions while structuring and formatting the figures, resulting in high variations across different papers. It is also challenging to extract information from charts amidst heavy clutter and deformation within the plot area. Even though the color is an essential cue for differentiating the plot data, it may only sometimes be present because many figures frequently reuse similar colors and some are even published in grayscale. Also, figure parsing presents an additional challenge because there is only one exemplar (the legend symbol) available for model learning, in contrast to natural image recognition tasks where the desired amount of labeled training data can be obtained to train models per category. Due to these challenges, there currently needs to be a system that can automatically parse data from scientific figures/charts.

In this paper, we make three key contributions. First, we propose ChartParser, a fully automated pipeline that leverages deep learning, OCR, and image processing techniques to extract all figures from a research paper, classify them into various chart categories and retrieve useful information from them, specifically bar charts. Second, we address some of the key challenges present in existing systems. For example, our system can parse legend and utilize color information for data association. It is also robust to variations in the figure designs and has no assumptions related to the position of axes, legend, etc. And finally, we demonstrate the viability of our approach by applying our pipeline to a real-world dataset of research papers from different sources.

\section{Related Work}
%

Chart understanding in scientific literature has recently gained much traction and there have been several attempts to classify charts using heuristics and expert rules. Various machine learning-based algorithms that rely on handcrafted features such as histogram of oriented gradients (HOG), scale-invariant feature transform (SIFT), and others have been proposed in the literature \cite{hog, sift}. Several deep learning algorithms for chart and table image classification have recently been introduced \cite{DL1, DL2}, and \cite{DL3}.

There is another line of work on interpreting text components in chart images \cite{TC1, TC2, TC3, TC4, TC6, TC5}. Although semi-automatic software solutions are available for data extraction from charts, using them requires the user to manually define the chart's coordinate system, provide metadata about the axes and data or click on the data points \cite{manual1, manual2, manual3}.

One of the difficulties in accurately parsing bar charts is dealing with different types of bar charts in scientific literature. Previous work, for example, \cite{heuristic1, heuristic2}, focused on developing heuristic models that detect key elements such as bars, legends, etc. Similarly, machine learning has also been used recently to detect chart components (e.g., bar or legend) \cite{detection1}. Also, a deep learning object detection model is trained in \cite{detection2} to identify sub-figures in compound figures. However, neither of these works extracted data values from bar charts. Using synthetic data produced by the matplotlib toolkit, \cite{matplotlib} created a model to boost the accuracy while parsing bar values.

Most of the previous methods do not parse the legend. Some assumed that the legend was always placed below the chart \cite{matplotlib} or horizontally along the same line \cite{legend}. This limits the applicability of these models. Previous work was mostly created for visualizations in grayscale, as they did not parse color information from the legend. Also, there has been less focus on measuring the accuracy of detecting the axes or label values. Quantifying the accuracy of obtaining this semantic information is essential for understanding the capping limits in this evaluation process. Even though the process of extracting information from charts and other infographics has been extensively explored, to our knowledge, prior work has several shortcomings as discussed above. As a result, we propose a fully automated system for data extraction from bar charts which solves these existing limitations and can be extended to other types of charts, including line charts, scatter plots, etc. 

\begin{figure*}[ht]
  \centering
   \includegraphics[width=0.9\linewidth, height=260pt]{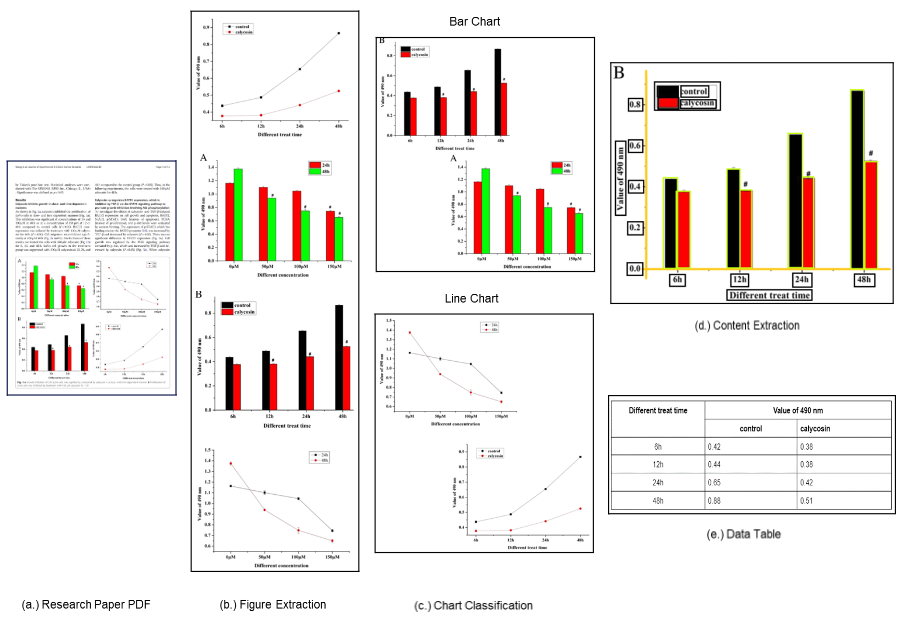}
   \caption{Illustration of the ChartParser pipeline}
   \label{fig:pipeline}
\end{figure*}

\section{Methodology}
\label{sec:methodology}
This section discusses our proposed pipeline to convert bar charts from scientific publications into data tables. The process is divided into three steps: First, we extract figures from research papers. Second, we detect bar charts from the extracted figures. And finally, we extract content from bar charts to obtain the desired data tables. These three steps are depicted in Figure \ref{fig:pipeline}.

\subsection{Figure Extraction}
To segment all the figures from a research paper, we use a pre-trained image segmentation model based on Mask R-CNN architecture from Detectron2 model zoo to decompose a document into five categories: title, text block, list, figure, and table. The model is based on the ResNet50 feature pyramid network (FPN) base config and is trained on the PubLayNet dataset for document layout analysis. 

\subsection{Figure Classification}
Most of the figures extracted are charts including tree diagrams, network diagrams, bubble charts, etc. This step describes the chart classification model employed to detect bar charts. 
 
\subsubsection{Chart Images Dataset} 
\label{dataset}
We create a chart dataset to train and evaluate our chart classification model. We use the Python module google images download to obtain charts from 13 categories (scatter plots, bar charts, line charts, etc.), 1000 images from each category. Then, we manually identify and remove some incorrect samples of downloaded charts. Finally, we obtain a ground-truth dataset of charts with a total of approximately 12k charts, including 978 bar charts.

\subsubsection{Chart Classification Model}
We try out different models pre-trained on the ImageNet dataset and fine-tune them on the figure dataset created. All the layers but the final convolutional layer were frozen. The fully-connected layer uses a softmax function to classify figures into 13 chart categories. Using Adadelta as the optimizer, we re-train the convolutional layer and the additional fully-connected layer for 30 iterations. We also add a dropout layer with a rate of 0.3 before the final fully-connected layer to avoid overfitting. Despite similar accuracy achieved by all the baselines, we choose MobileNet as it uses far less parameters on ImageNet than others. 

\subsection{Content Extraction}
Content Extraction from charts is a complex process and in this step, we employ OCR and image processing techniques to extract relevant content from bar charts through various modules. 

\subsubsection{Axes Detection}
We convert the image into a binary one, and then, obtain the max-continuous ones along each row and column. For this, we scan the matrix vertically and horizontally to trace the continuity of black pixels within the adjacent columns and rows. Finally, the y-axis is the first column where the max-continuous 1s fall in the region [max - threshold, max + threshold], where a predetermined threshold (=10) is assumed. Similarly, for the x-axis, the last row is chosen based on where the maximum continuous 1s fall within the range [max - threshold, max + threshold].

\subsubsection{Text Detection}
We apply Azure Cognitive Service (ACS) Optical Character Recognition (OCR) to detect text within a chart and extract all the rectangular bounding boxes of the detected text.

\subsubsection{Axes Ticks Detection}
We filter all the text boxes below the x-axis and to the right of the y-axis. Further, we run a sweeping line from the x-axis to the bottom of the image and the line which intersects with the maximum number of text boxes provides the bounding boxes for all the x-axis ticks. A similar algorithm is used for detecting y-axis ticks using a vertical sweeping line.

\subsubsection{Axes Label Detection}
We filter the text boxes present below the x-axis ticks and again, run a sweeping line from the x-axis ticks to the bottom of the image. While doing so, the line intersecting with the maximum number of text boxes provides us with all the bounding boxes for the x-axis label. Similarly, we also obtain the y-axis label using a vertical sweeping line.

\subsubsection{Legend Detection}
Firstly, we remove the axes labels and ticks bounding boxes. Then, we also remove boxes containing only a single "I" character because these are typically read as error bars and finally, we also remove text boxes with numeric values placed above bars. This implies that only legend names and color boxes are found in the remaining text boxes. We combine the bounding boxes with distances under 10px into a single legend name because the legend names might have multiple words. We organize these bounding boxes into groups where each member is either horizontally or vertically aligned with at least one other member. Finally, the maximum length group gives the bounding boxes for all the legends.

\subsubsection{Legend Color Estimation}
The color boxes are assumed to be on the left or right side, depending on the placement of text bounding box within the legend extracted in the previous module. Pixels within a box should ideally all have the same pixel value. Since, these values could change for several reasons (such as image compression, scanning, etc.), we start a new group with a random pixel and gradually add pixels whose R, G, and B values are no higher than 5 compared to the average of all the pixels in the group. The color of a legend label is determined by taking the average of all the pixels in the largest group of the R, G, and B channels. Later, bars matching a specific legend are identified using these colors.

\subsubsection{Data Extraction}
The bounding boxes for each legend are whitened, and we eliminate all the white pixels from the original chart image. The colors decided upon in the previous module serve as the initial clusters as all of the image's pixel values are further divided into clusters. Then, we divide the given plot into multiple plots, one for each cluster. In other words, by clustering, we break down a stacked bar chart into several simpler plots. Then, we obtain all contours within the plot and subsequently, pick the closest bounding rectangle for each label. Further, we require a mapping function to map pixel values to actual values in the chart. Hence, we use the value-tick ratio  (\(\alpha\)) to estimate the height of each bar. To find this ratio, we divide the average of the actual y-label ticks ($ N_{ticks} $) by the average distance between ticks in pixels ($ \Delta d $).  
\begin{equation}
    \alpha = N_{ticks}/ \Delta d
\end{equation}
Finally, the bar chart's y values are defined as y value = \(\alpha\) \(\times\) H, where H is the bar's height. After getting all the relevant information, we create a data table using the same as shown in Figure \ref{fig:pipeline} (e.).

\section{Results}
This section focuses on creating a test dataset of bar charts from research papers and evaluating various components of our pipeline on this dataset to demonstrate the viability of our approach.

\begin{table}
  \centering
  \begin{tabular}{c|c|c}

    Component & Accuracy (\%)  \\

    X-axis & 98 \\
    Y-axis & 96  \\
    X-axis label & 98 \\
    Y-axis label & 96 \\
    X-axis ticks & 89 \\
    Y-axis ticks & 84 \\
    Legend & 92 \\
    Legend color & 87 \\
    Data association & 76 \\

  \end{tabular}
  \caption{Content extraction accuracy}
  \label{tab:table1}
\end{table}

\subsection{Test Dataset}
We sample research papers from two data sources: arXiv and PubLayNet. From the arXiv dataset published on Kaggle \cite{kaggle}, we obtained research paper PDFs published in the years 2019-2021 and the resulting dataset consisted of around 10,024 papers. Also, we use a subset of the PubLayNet dataset \cite{publaynet} and obtain approximately 15k document images from the same. Then, we apply the first two steps of our fully automated pipeline to these research papers, as mentioned in the previous section\ref{sec:methodology}. First, we extract approximately 51k figures from the research papers dataset using our image segmentation model, and then, on applying our chart classification model to these figures, we obtain approximately 2,112 bar charts. To evaluate our system, we sample 100 bar charts and manually annotate the relevant data, including axes, axes label, axes tick's values, legend, legend color, and the textual bounding boxes. 

\subsection{Chart Classification}
The accuracy of our chart classification model is calculated using stratified five-fold cross validation. Here, we use 20\% of the chart images dataset, created using google images download API, as our validation set and the category wise performance (average accuracy) of our model is presented in Table \ref{table2}. We observe that for bar charts, our model achieves an accuracy of 97.8\%.

\begin{table}
  \centering
  \begin{tabular}{c|c|c}
    Category & Accuracy (\%)  \\
    Bar Chart & 97.8 \\
    Line Chart & 96.86  \\
    Scatter Plot & 92.00 \\
    Pareto chart & 84.20 \\
    Pie Chart & 91.52 \\
    Venn Diagram & 87.88 \\
    Box Plot & 94.56 \\
    Network Diagram & 68.97 \\
    Map & 79.26 \\
    Tree Diagram & 69.09 \\
    Area Graph & 88.00 \\
    Flow Chart & 75.54 \\
    Bubble Chart & 92.20 \\
  \end{tabular}
  \caption{Category wise average accuracy of the chart classification model}
  \label{table2}
\end{table}

\subsection{Text Recognition}
We use the Intersection Over Union (IoU) metric to assess our text detection module. This metric determines the bounding boxes that most closely match the predicted and actual ones, calculates the area of the intersecting region divided by the area of the union region for each match, and considers the prediction successful if the IoU measure is higher than the threshold, for example, 0.5.We achieve an F1-score of 0.935 with an IoU threshold of 0.5 and this demonstrates that our module detects text bounding boxes within the plot area fairly well.

\subsection{Content Extraction}
The performance of the final content extraction process depends on the sequential performance of each module, i.e., axis detection, axis tick values extraction, label extraction, legend detection, and so on. First, we apply the OCR and image processing techniques to the test dataset and extract relevant content. Then, we compare the outcome with the manually annotated data and obtain module-wise evaluation metrics presented in Table \ref{tab:table1}.

\section{Limitations and Future Work}
This section mentions the existing limitations of our fully-automated pipeline and also proposes future works for improvement.

Currently, there is a problem with our proposed pipeline that prevents it from successfully parsing the plotted data when there is a lot of clutter. We can employ vascular tracking methods like those described in \cite{clutter} to solve this. 

Also, our pipeline fails to recognize axes when there is no solid line indicating the y-axis. In this scenario, the y-axis can be identified by recognizing bounding boxes along a vertical line in the bar chart. Also, when the x-axis is at the top of the graphic, x-axis detection may fail. This case can be handled by employing a bidirectional sweeping line with heuristic rules.

We also realize that the axes, legend, and data extraction modules are currently modeled and trained independently in our figure analysis approach. It can be an exciting approach to jointly model and train them together within an end-to-end deep network.

In our future work, we will extend our pipeline to other types of charts as well including line charts, scatter plots, etc. which have an L-shaped axis, similar to bar charts and also, follow a similar algorithm for extraction of chart elements such as axes, labels, ticks, legends, etc. Instead of simply presenting the raw data in tabular form, we can also generate insights from the data by employing reasoning on chart images at a high level by finding relationships between various chart elements.

\section{Conclusion}
In this paper, we present our ongoing work in making scientific documents accessible to the blind, low-vision, and print-disabled individuals. Our work focuses on the problem of poor accessibility of infographics/charts in research papers. We propose an end-to-end pipeline to extract all figures from a research paper, classify them into various chart categories, obtain relevant information from them, specifically bar charts and present the retrieved content into accessible data tables. Finally, we apply our pipeline to a test dataset of research papers from two different sources: arXiv and PMC to demonstrate the viability of our approach. We continue to work towards making charts fully accessible to print-impaired individuals by overcoming the existing limitations of our work.

\bibliography{aaai22}

\end{document}